\pdfoutput=1
\documentclass[11pt]{article}

\usepackage[final]{acl}
\usepackage{multirow}
\usepackage{makecell}
\usepackage{graphicx} 
\usepackage{ltablex}
\usepackage{booktabs}
\usepackage{times}
\usepackage{latexsym}
\usepackage{hyperref}
\usepackage{xcolor}
\usepackage{ragged2e}
\keepXColumns
\usepackage{listings}
\usepackage{minted}
\usepackage[most]{tcolorbox}

\lstdefinelanguage{json}{
    basicstyle=\ttfamily\small,
    numbers=none,
    showstringspaces=false,
    breaklines=true,
    frame=none,
    literate=
     *{0}{{{\color{blue}0}}}{1}
      {1}{{{\color{blue}1}}}{1}
      {2}{{{\color{blue}2}}}{1}
      {3}{{{\color{blue}3}}}{1}
      {4}{{{\color{blue}4}}}{1}
      {5}{{{\color{blue}5}}}{1}
      {6}{{{\color{blue}6}}}{1}
      {7}{{{\color{blue}7}}}{1}
      {8}{{{\color{blue}8}}}{1}
      {9}{{{\color{blue}9}}}{1}
      {:}{{{\color{black}{:}}}}{1}
      {,}{{{\color{black}{,}}}}{1}
      {"}{{{\color{gray}{"}}}}{1}
}

\tcbset{
  colback=gray!5,
  colframe=gray!50!black,
  left=2mm,
  sharp corners,
  boxrule=0.4pt
}

\usepackage[LGR,T1]{fontenc}
\usepackage{enumitem}
\usepackage[utf8]{inputenc}
\usepackage[greek,english]{babel}

\usepackage{microtype}

\usepackage{inconsolata}

\usepackage{graphicx}

\title{Explain the Flag: Contextualizing Hate Speech Beyond Censorship}

\author{
  \textbf{Jason Liartis\textsuperscript{1,2}},
  \textbf{Eirini Kaldeli\textsuperscript{1,2}},
  \textbf{Lambrini Gyftokosta\textsuperscript{3}}
\\
  \textbf{Eleftherios Chelioudakis\textsuperscript{4,5}},
  \textbf{Orfeas Menis Mastromichalakis\textsuperscript{1,6}}
\\
\\
  \textsuperscript{1}National Technical University of Athens,
  \textsuperscript{2}Datoptron,
  \textsuperscript{3}Independent Researcher
\\
  \textsuperscript{4}Homo Digitalis,
  \textsuperscript{5}University of the Aegean, 
  \textsuperscript{6}Instituto de Telecomunicações
}

\begin{document}
\maketitle
\begin{abstract}

% ------- NEW ABSTRACT START -------
Hate, derogatory, and offensive speech remains a persistent challenge in online platforms and public discourse. While automated detection systems are widely used, most focus on censorship or removal, raising concerns for transparency and freedom of expression, and limiting opportunities to explain why content is harmful. To address these issues, explanatory approaches have emerged as a promising solution, aiming to make hate speech detection more transparent, accountable, and informative. In this paper, we present a hybrid approach that combines Large Language Models (LLMs) with three newly created and curated vocabularies to detect and explain hate speech in English, French, and Greek. Our system captures both inherently derogatory expressions tied to identity characteristics and direct group-targeted content through two complementary pipelines: one that detects and disambiguates problematic terms using the curated vocabularies, and one that leverages LLMs as context-aware evaluators of group-targeting content. The outputs are fused into grounded explanations that clarify why content is flagged. Human evaluation shows that our hybrid approach is accurate, with high-quality explanations, outperforming LLM-only baselines. Our source code is available at \url{https://github.com/ails-lab/detoex}.
\end{abstract}

\section{Introduction}

% \textcolor{red}{Warning: This paper contains terms and expressions that may be offensive or disturbing to some readers. They are included solely for illustration, and are not intended to offend or endorse such language.} 
\textcolor{red}{Warning: This paper contains terms and expressions that may be offensive or disturbing to some readers. They are included solely for illustration and are not intended to endorse such language.} 

The spread of hate speech online has become a pressing concern with serious personal, social, and legal consequences. Beyond harming individuals, it can escalate social tensions and, in severe cases, contribute to discrimination, hostility, and violence. These risks highlight the need to monitor and address harmful language for the sake of well-being, social cohesion, and legal responsibility. Institutions have recognized this urgency: the European Union has launched initiatives to analyze, regulate, and counteract online hate speech \cite{hatelab2023}\footnote{\url{https://europeanonlinehatelab.com/}}, and the United Nations has developed a dedicated strategy and plan of action to identify, prevent and confront hate speech\footnote{\url{https://www.un.org/en/hate-speech/un-strategy-and-plan-of-action-on-hate-speech}}.
% , acknowledging its potential to incite violence and undermine social unity \footnote{\url{https://www.un.org/en/hate-speech/un-strategy-and-plan-of-action-on-hate-speech}}.

%Following established international efforts, 
% In line with the widely-accepted definitions used in the 
In line with existing literature~\cite{tonneau2024}, this work adopts the United Nations' definition of hate speech. According to the UN, hate speech is \emph{``any kind of communication in speech, writing or behaviour, that attacks or uses pejorative or discriminatory language with reference to a person or a group on the basis of who they are, in other words, based on their religion, ethnicity, nationality, race, colour, descent, gender or other identity factor''}\footnote{\url{https://www.un.org/en/hate-speech/understanding-hate-speech/what-is-hate-speech}}. This interpretation is broader than purely legal definitions, which often restrict hate speech to explicit incitement to violence. It covers a wide spectrum of harmful communication, from derogatory expressions and stereotypes to insults and threats, as long as they are rooted in identity-based characteristics.
Personal insults or generic offensive language are generally not included, unless they target individuals or groups based on such characteristics. Hate speech can thus emerge in two ways: through inherently derogatory expressions tied to identity, or through content targeting individuals or groups based on their identity, even without explicit profanity.
% Addressing these multiple manifestations is essential for systems that aim to capture the full scope of harmful communication.

Current automated systems for moderating hate speech typically focus on flagging or removing harmful content, often without providing users with any justification. While such interventions may reduce exposure to offensive material, they also introduce two significant risks: users are not given the opportunity to understand how their language may perpetuate stereotypes or cause harm, and moderation actions may appear arbitrary or biased. As a result, many users encounter content removals or labels without explanation, leaving them unable to interpret or contest these decisions. In response to these shortcomings, recent research has begun advocating for explanatory approaches that contextualize offensive and derogatory language rather than simply removing it \cite{epstein2022,menis-mastromichalakis-etal-2025-dont}.

In this work, we present a hybrid approach designed to capture the full scope of hate speech while providing grounded, contextualized explanations. A central contribution of our study is a set of curated vocabularies in English, French, and Greek containing terms that are inherently offensive or derogatory toward specific groups. These vocabularies were constructed by collecting and reviewing entries from Wiktionary\footnote{\url{https://www.wiktionary.org/}}
, including information about each term’s meaning in contentious and non-contentious contexts, as well as the identity characteristic(s) it targets. This curated resource provides reliable grounding for the LLM, improving both detection accuracy and explanation quality while ensuring coverage of evolving or rare expressions that LLMs may fail to recognize.
Our system integrates two complementary pipelines: a term-based pipeline that uses lemmatization and string matching to identify potentially problematic terms and leverages an LLM to disambiguate their meaning in context; and a term-free pipeline in which an LLM detects content explicitly targeting individuals or groups based on identity characteristics. The outputs of both pipelines are fused by an LLM to generate grounded explanations that clarify why specific content is flagged.
We evaluate our approach through a human study on English, French, and Greek texts, assessing both detection performance and explanation quality, and show that our hybrid system outperforms LLM-only baselines.

\section{Related Work}
\label{sec:rel-work}

Online abusive and harmful language became a prominent challenge with the rise of early internet communities, where moderation was handled manually by platform administrators, forum moderators, and community managers. As online platforms and social media networks grew, the volume and velocity of user-generated content made manual moderation unsustainable, motivating the development of automated approaches. Early systems relied on traditional machine learning models such as SVMs \cite{warner-hirschberg-2012-detecting}, often combined with curated dictionaries \cite{tulkens2016dictionary} or ensemble models \cite{burnap2014hate}. With the emergence of deep learning, convolutional and recurrent neural networks (CNNs, LSTMs) were applied to classify abusive or hateful content \cite{del2017hate, mathur-etal-2018-detecting, meyer-gamback-2019-platform, chakrabarty-etal-2019-pay, modha-etal-2018-filtering}, followed more recently by transformer-based architectures and large language models (LLMs), which currently set the state of the art in detecting toxic language \cite{elmadany-etal-2020-leveraging, alonso2020hate, davidson-etal-2020-developing, yao2024personalised, plaza2023respectful, vargas2026selfexplaininghatespeechdetection}.

Alongside methodological advances, many studies have focused on developing resources to support hate speech detection and contextualization. Structured frameworks for classifying offensive content provide a basis for systematic analysis \cite{banko-etal-2020-unified, kurrek-etal-2020-towards}, while lexicons, vocabularies, and annotated datasets offer collections of terms associated with abusive or derogatory speech for training and evaluation \cite{tonneau2024, elsherief2021, sap2020}. Building on this line of work, we created curated vocabularies in English, French, and Greek containing terms inherently offensive or degrading toward identity-based groups. Unlike many existing lexicons that list only offensive terms, our vocabularies include the meaning and nuances of each term in both contentious and non-contentious contexts, along with the identity characteristic it targets (e.g., religion, sexual orientation), providing a foundation for grounded, context-aware analysis.

Despite these resources, automated detection remains challenging due to contextual and linguistic nuances. Many data-driven models exhibit biases \cite{wiegand2019detection, davidson-etal-2019-racial, xia-etal-2020-demoting, zhang-etal-2020-demographics, sap-etal-2019-risk} and robustness issues \cite{Kaushik2020Learning, sen-etal-2022-counterfactually, korre2023harmful}, which limit fairness and applicability in real-world settings. Incorporating context is essential to reduce false positives and improve detection quality \cite{kennedy-etal-2020-contextualizing, bourgeade-etal-2024-humans}, an area where LLMs have shown particular promise.

Beyond detection accuracy, there is growing interest in approaches that justify and contextualize hate speech moderation. Traditional flagging or removal of content, while reducing exposure to offensive material, often provides users with no rationale, limiting understanding and raising concerns about unfair censorship. Explanatory approaches aim to bridge this gap, helping users recognize and avoid offensive language. Recent works employ LLMs and curated rsources to generate rationales or highlight problematic segments \cite{epstein2022, nirmal2024, yang2023, fan2023, 10.1007/978-3-031-88711-6_24, menis-mastromichalakis-etal-2025-dont}.
However, most such approaches focus on English and other high-resource languages, with lower-resource languages such as Greek receiving less attention. While automated approaches often perform well in high-resource languages, they may fall short in lower-resource contexts. Creating resources for these languages, such as the curated vocabulary presented here, can boost detection accuracy and explanation quality.
% By combining curated vocabularies with LLMs, our method detects and explains hate speech that manifests both through inherently derogatory terms and through content targeting groups based on identity characteristics. This hybrid approach produces grounded explanations informed by both the vocabularies and textual context, enhancing interpretability, accountability, and applicability across languages.

\section{Methodology}

\subsection{Aspects of Hate Speech}
\label{sec:aspects}

% The conceptualisation of hate speech depends on the characteristics of social groups with which the targets of harmful language are associated. For our analysis, we consider the following group characteristics by putting together identity-related dimensions considered in online datasets and literature~\cite{elsherief2018,ousidhoum2019}:  Gender; Sexual Orientation; Race; Ethnicity; Religion; Political Affiliation; Socioeconomic; Age; Disability; Addiction; and Physical Appearance.
The conceptualization of hate speech relies on the identity characteristics of the individuals or groups being targeted. For our analysis, we consider the following characteristics, based on prior research and online datasets~\cite{elsherief2018,ousidhoum2019}: \emph{gender, sexual orientation, race, ethnicity, religion, political affiliation, socioeconomic status, occupation, age, disability, addiction,} and \emph{physical appearance}. 

Drawing on these identity characteristics, hate speech manifests in two main ways. One relies on inherently derogatory terms that carry offensive meaning on their own, while the other involves hostile or demeaning expressions directed at individuals or groups specifically because of their identity, even when no explicit slur is present. Capturing both manifestations is essential to understanding the full spectrum of harmful language, including explicit and implicit expressions. We describe each in detail below, providing examples and highlighting the role of context in determining harmfulness.

\emph{Use of inherently derogatory terms:} This refers to words or expressions whose meaning is offensive or degrading toward a specific group. Such terms include slurs and insults that are inherently tied to an identity characteristic, even when not directed at a member of that group. For example, the Greek term ``\textgreek{αδελφή}'', when used with the meaning of ``sissy'' or ``faggot'', is derogatory toward gay people. It can be used as a direct insult against members of the queer community, but it can also appear in contexts where it is not aimed at a person because of their sexual orientation (e.g., saying ``don't be such a sissy'' to a friend). Even in these cases, the term perpetuates stereotypes about gay people and is therefore considered derogatory. Context is crucial because some words have neutral meanings (e.g., ``\textgreek{αδελφή}'' also means ``sister'' or ``nun'') but also reclamation by targeted communities can change how they are perceived and used in different environments.

%\emph{Targeted language against groups or individuals based on identity:} This refers to expressions that explicitly attack a group or an individual because of their membership in that group. Here, the harmfulness does not necessarily come from the use of a dedicated slur, but from the way otherwise common insults or statements are applied to identity characteristics. For example, a statement like ``He should be arrested and jailed; he's a Muslim, so he's a terrorist and a danger to public safety, just like the others of his kind'' targets an individual based on their religion, perpetuating a harmful stereotype that all Muslims are terrorists. 
%In many cases, we see a combination of both forms of hate speech manifestation. For instance, calling a woman a ``bitch'' in a sentence such as ``That bitch is always complaining, why can't women be more rational like us?'' both relies on an inherently derogatory term and targets her based on gender. Similarly, saying ``That nigga always causes trouble, like the rest of them.'', uses an inherently offensive term directed at a black individual while perpetuating a racial stereotype. 

\emph{Language directed against groups or individuals based on their identity:} This refers to expressions that explicitly attack a group or an individual because of their identity, even without inherently derogatory terms. For example, ``He should be arrested and jailed; he's a Muslim, so he's a terrorist and a danger to public safety'' targets an individual based on their religion, perpetuating a harmful stereotype that all Muslims are terrorists. 

Often, both forms occur together. Calling a woman a ``bitch'' in ``That bitch is always complaining, why can't women be more rational like us?'' uses an inherently derogatory term while targeting her gender. Similarly, ``That nigga always causes trouble, like the rest of them.'', combines an inherently offensive term with a racial stereotype.

\subsection{The Vocabularies}

To address hate speech in the form of inherently derogatory terms, we constructed three dedicated vocabularies, one for each language in our study: English, French, and Greek. Each vocabulary contains terms and expressions that are, by themselves, derogatory or offensive toward a group or its members, providing a systematic basis for detection. Each entry includes: (1) the term; (2) a description covering all meanings (both offensive and non-offensive); (3) the identity characteristic(s) targeted; and (4) a link to the source repository. A detailed example is provided in Appendix~\ref{app:voc_example}.

We selected Wiktionary as the source for our vocabularies, as it provides broad coverage across languages and has been shown to be ``\emph{a reliable and linguistically rich resource whose collaboratively constructed entries show a quality largely comparable to expert-made dictionaries}'' \cite{meyer-gurevych-2012-exhibit}. Its open availability ensures transparency and reproducibility, while its active community continuously updates entries, keeping the resource current with evolving language. Wiktionary also provides structured metadata, including usage labels such as ``derogatory'', ``offensive'', and ``vulgarities'', which can facilitate the systematic identification of inherently derogatory terms. 

Our methodology for building the vocabularies involved five steps: \\
\textbf{1. Initial collection:} Retrieved Wiktionary terms tagged with relevant labels via the Wiktionary API, parsing and cleaning the returned HTML. This yielded 11,310 English, 3,749 French, and 965 Greek terms. See Appendix~\ref{app:voc_example} for more details. \\
\textbf{2.  Filtering:} Retrieved terms were filtered to remove common slurs and terms that are not inherently derogatory toward a group. For Greek, filtering was performed manually by human experts, while for English and French we used LLM-assisted filtering with Claude Sonnet 3.7\footnote{claude-3-7-sonnet-20250219-v1:0}, with human supervision and sampling validation. \\
\textbf{3. Categorization:} Tagged filtered terms with the identity characteristics they target. Manual categorization was performed for Greek, while English and French used LLM-assisted categorization with human supervision and sampling validation. \\
\textbf{4. Generation of enriched description:} All Wiktionary descriptions for each filtered term were fed to Claude Sonnet 3.7 with an appropriate prompt to generate continuous text that describes the meaning(s) of the term with an emphasis on why and under which circumstances the term is used in an offensive way (detailed prompts in Appendix~\ref{app:prompts}). \\
\textbf{5. Human Validation:} An expert team of 2 legal professionals\footnote{The experts hold a Master of Laws degree and work on matters related to hate speech, online abusive speech, anti-discrimination law, and AI-related regulatory and governance issues.} reviewed and corrected, where needed, the LLM outputs from steps 2–4. The Greek vocabulary was fully validated, while for English and French a representative sample (10\% of entries) was reviewed. The experts found results satisfactory in over 90\% of cases; in most of the remaining cases, the outputs were not wrong but required minor corrections.

% Key challenges included addressing semantic shifts due to community reclamation of previously derogatory terms, which required prompt adjustments and manual corrections to prevent over-flagging by downstream systems. 
The resulting vocabularies contain 3,904 English, 1,644 French, and 288 Greek entries. See Appendix~\ref{app:vocab-stat} for the complete breakdown by category and language. The vocabularies are provided in the supplementary materials and will be made openly accessible upon publication.

\subsection{System Architecture}\label{sec:system_arch}

\begin{figure*}[t]
  \centering
  \includegraphics[width=0.85\textwidth]{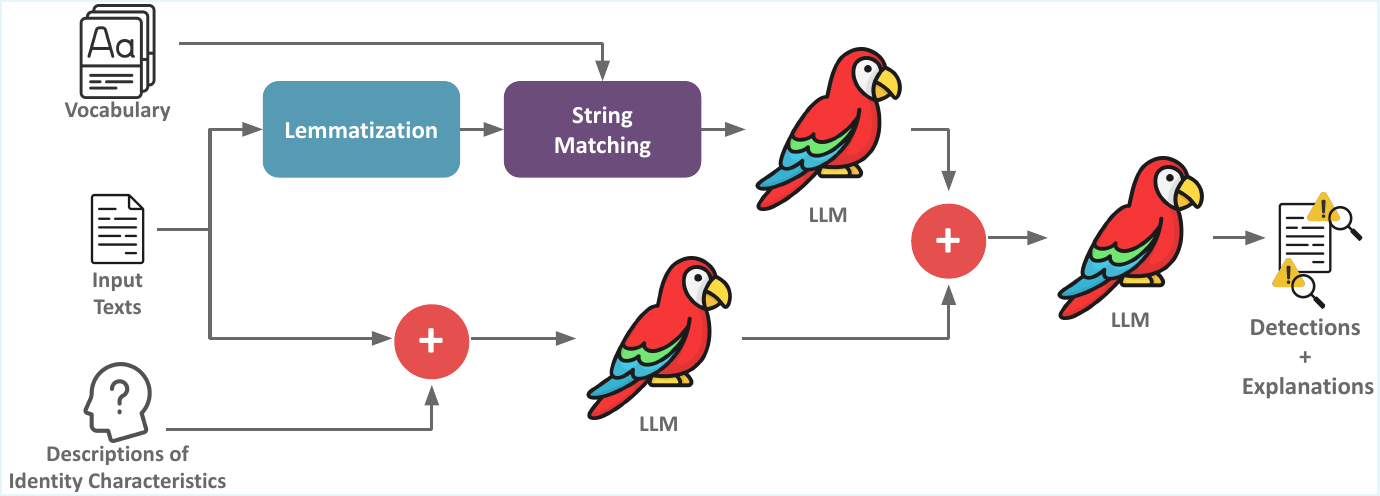}
  \caption{Overall system architecture illustrating the two pipelines for hate speech detection and explanation.}
  \label{fig:system-architecture}
\end{figure*}

Our detection system implements two complementary objectives through parallel processing pipelines, as shown in Figure~\ref{fig:system-architecture}.
The first pipeline, which we refer to as the \emph{term-based} pipeline, detects inherently derogatory terms using the curated vocabularies. It combines string matching with LLM-powered semantic disambiguation. Terms are detected by matching lemmatized forms of the terms in the input text. Advanced matching logic handles multiple matches (retaining the longest) and grammatical variations sharing the same lemma (retaining the shortest Levenshtein distance). An LLM then resolves semantic ambiguity, determining whether the detected term is used contentiously. The LLM receives the source text and the vocabulary description for each term and outputs a decision along with a free-text explanation of why the term is offensive in context.

The second pipeline, which we refer to as the \emph{term-free} pipeline, detects hate speech targeting individuals or groups without relying on specific vocabulary terms. It leverages an LLM grounded in curated identity characteristics to decide whether a text contains hate speech directed at individuals or groups based on their identity or beliefs, including slurs and other expressions not captured in the vocabularies. The LLM receives the source text, the identity characteristics, and dedicated instructions. We consider this pipeline as the system's baseline as it mirrors how many LLM detections systems operate, using a system prompt that describes the task without augmenting it with an external lexical resource.

A text is classified as free of hate speech only if both pipelines agree. If one pipeline detects hate speech, the text is flagged and the corresponding output is returned. If both pipelines flag it, an LLM fuses the outputs, removing redundancy and producing a coherent, unified explanation.
% \end{enumerate}
%(1) term-based detection of vocabulary-defined politically-charged  terms with contextual disambiguation, and (2) detection of  language directed against specific demographic groups, including expressions not captured in our curated vocabularies.

%\paragraph{Pipeline B: Group-Targeted Detection}
%This pipeline analyzes text for any  expressions (including common slurs) explicitly targeting individuals or groups based on protected characteristics (religion, ethnicity, sexual orientation, etc.) and relies exclusively on an LLM.
%The model receives our definition of politcally-charged  language along with the text and dedicated instructions to identify offensive language directed at demographic groups, producing explanations for detected .

% The two complementary pipelines are applied in parallel.
%\paragraph{Output Fusion}
%The pipelines operate in parallel.
% Single-pipeline detections are outputed as they are, while dual detections undergo fusion by an LLM to eliminate redundancy and create coherent unified outputs.

%\paragraph{Model Selection}
For comparative evaluation, we deployed two LLM configurations per language: Claude Sonnet 3.7 as the proprietary large model, and Llama-family variants as smaller open-weight alternatives. Language-specific LLMs included Llama-Krikri-8B-Instruct~\footnote{https://huggingface.co/ilsp/Llama-Krikri-8B-Instruct}
for Greek and Hermes 3 Llama 3.1 8B~\footnote{https://huggingface.co/NousResearch/Hermes-3-Llama-3.1-8B}
for English and French (selected after standard Llama-3.1-8B exhibited excessive guardrail restrictions).
The models from the Llama family were selected as a low-resource open-weight option as they can be deployed on a commercial GPU with at least 24GB of VRAM, while Claude Sonnet 3.7 was selected to test whether a sufficiently large model would implicitly internalize the kinds of contextual knowledge that our vocabulary explicitly provides.
 
Lemmatization employed Stanford Stanza \cite{qi2020stanza} for English and French, and both Stanza and the ILSP lemmatizer \cite{prokopidis2020neural} for Greek, which has demonstrated superior performance for Greek terms.
% Due to deployment constraints with ILSP's API-only access, the production system defaults to Stanza, though offline evaluation used ILSP processing.
%Implementation details and performance comparisons are provided in Appendix ??.

%Different prompt variations (e.g. breaking down the task to simpler steps, asking the model to assign a  rating, dedicated instructions to avoid over-detection etc) have been tested on sample data from the evaluation datasets (see Section 1.2.4) until achieving satisfactory results in terms of decision and explanation quality. 
%\paragraph{Prompt Engineering}
Initial prompts were refined through qualitative evaluation of early outputs. Key enhancements included chain-of-thought formatting for term disambiguation, explicit instructions for reclaimed language, and guidelines for distinguishing direct from indirect speech and quotations (treated as non-hateful). While these refinements improved performance and explanation clarity, extensive prompt engineering or comparison of alternative prompting strategies was beyond the scope of this work.
Cross-language evaluation showed no significant quality differences between prompts written in the target language and those with English instructions.
See Appendix~\ref{app:prompts} for English prompts.

\section{Experiments}

% ================ NEW ================
To assess the effectiveness of our system, we conduct a series of experiments that evaluate both its detection performance and the quality of the explanations it produces. Beyond assessing overall system behaviour, we also examine how the different components of our approach contribute to these results. All experiments are conducted on a multilingual dataset that we constructed through manual annotation.

\subsection{Evaluation Dataset}
For our evaluation, we required real online texts that contain hate speech according to the definition adopted in this work, while also ensuring comparability across English, Greek, and French. To guide dataset selection, we established four criteria. First, the texts should be authentic user-generated content. Second, they should include cases of identity-targeting hate speech rather than only general toxicity or personal insults. Third, the datasets for the three languages should contain similar types of texts to maintain a consistent evaluation setup. Fourth, the selected texts should provide wide coverage of different identity characteristics, so that the evaluation reflects the diverse ways hate speech can manifest.

While several resources exist for English and French, Greek datasets are limited. Moreover, many available datasets across all three languages focus on offensive or toxic language more broadly, meaning they often include profanity or personal insults that do not meet our definition of hate speech. Some are also topic-specific (for example, restricted to sexism or racism), which limits coverage of the identity characteristics.

Taking these considerations into account, we selected the Greek subset of OffensEval2020 \cite{zampieri-etal-2020-semeval}, 
the English Hate Speech Superset, 
and the French Hate Speech Superset \cite{tonneau2024} 
as our source datasets. Because the Greek dataset contains only tweets, we also restricted the English and French selections to tweets to maintain consistency across languages. These datasets provide suitable material containing both hateful and non-hateful content spanning a range of topics and target groups.
It needs to be noted that these tweets do not contain user metadata such as gender, race, etc. which can influence the presence of hate speech, as in some edge cases slurs are present but used in a self-referential reclamatory way. Since these metadata are often not available in real world scenarios, we consider the pure-text setting to have wider applicability, and in practice contextual linguistic cues like tone, target, syntactic framing, and explicit references are often sufficient to distinguish reclaimed from derogatory use.
Moreover, the original labels do not always align with our definition of hate speech (see Appendix \ref{app:dataset-statistics}). Consequently, the original labels could not be used as ground truth.

To obtain reliable evaluation data, we selected and manually annotated 1,600 texts from the source datasets described above, 600 for English, 400 for French, and 600 for Greek. The selection ensured wide coverage of all identity characteristics and a roughly balanced distribution of hateful and non-hateful examples. This was achieved by considering the topic-related and toxicity labels in the original datasets to sample diverse and representative items across the three languages.

For the manual annotation, we recruited 18 annotators, all members or collaborators of organizations working on topics related to hate speech, digital rights, cyberbullying, and other online safety issues. Each text in the evaluation dataset was annotated independently by three annotators, who assigned one of three labels: \textit{Yes} (contains hate speech), \textit{Unsure}, or \textit{No} (does not contain hate speech). Details on the inter-rater agreement between annotators are provided in Appendix~\ref{app:inter-rater}.

Based on the annotators' responses, we created our main evaluation dataset using majority voting. Since each text was annotated by multiple individuals, we assigned a final label when the annotators leaned in one direction (e.g. two \textit{Yes} and one \textit{No}, or one \textit{Yes} and two \textit{Unsure}), discarding only the fully divergent cases (e.g., one \textit{Yes}, one \textit{No}, and one \textit{Unsure}).
To gain further insights into the behavior of our system, such as whether it tends to be more ``strict'' or more ``permissive'', we also constructed three additional variants of the evaluation dataset. These variants differ in how they resolve borderline cases, defined as instances in which at least one annotator selected \textit{Unsure} or where disagreement occurred. The three variants are:\\
\noindent-- \emph{Safe}: ignores all borderline cases.\\
\noindent-- \emph{Permissive}: if no evaluator has identified the text as positive, it is considered negative; otherwise it is considered positive. This favors treating borderline cases as non--hate speech.\\
\noindent-- \emph{Strict}: if no evaluator has identified the text as negative, it is considered positive; otherwise it is considered negative. This favors classifying borderline cases as hate speech.\\
See Appendix~\ref{app:dataset-statistics} for additional statistics on the constructed evaluation datasets, including class distributions and their (dis)agreement with the original dataset labels.

\subsection{Accuracy of Detection}
\label{sec:accuracy}

\begin{table*}[ht]
\centering
\resizebox{\textwidth}{!}{%
\begin{tabular}{cc
>{\centering\arraybackslash}m{1.2cm}
>{\centering\arraybackslash}m{1.2cm}
>{\centering\arraybackslash}m{1.2cm}
>{\centering\arraybackslash}m{1.2cm}
>{\centering\arraybackslash}m{1.2cm}
>{\centering\arraybackslash}m{1.2cm}
>{\centering\arraybackslash}m{1.2cm}
>{\centering\arraybackslash}m{1.2cm}
>{\centering\arraybackslash}m{1.2cm}
>{\centering\arraybackslash}m{1.2cm}
>{\centering\arraybackslash}m{1.2cm}
>{\centering\arraybackslash}m{1.2cm}
}
\toprule
& & \multicolumn{3}{c}{\makecell{Safe}} & \multicolumn{3}{c}{\makecell{Majority}} & \multicolumn{3}{c}{\makecell{Permissive}} & \multicolumn{3}{c}{\makecell{Strict}} \\
\cmidrule(lr){3-5} \cmidrule(lr){6-8} \cmidrule(lr){9-11}
\cmidrule(lr){12-14}
Lang & Model & 
Precision & Recall & \mbox{F1-Score} & Precision & Recall & \mbox{F1-Score} & 
Precision & Recall & \mbox{F1-Score} & 
Precision & Recall & \mbox{F1-Score} \\
\midrule
\multirow{2}{*}{EN} 
& \makecell{Claude}      & \textbf{0.97} & \textbf{0.90} & \textbf{0.93} & \textbf{0.92} & \textbf{0.89} & \textbf{0.90} & \textbf{0.47} & \textbf{0.90} & \textbf{0.62} & \textbf{0.98} & \textbf{0.78} & \textbf{0.87} \\
& \makecell{Llama-based} & 0.84 & 0.80 & 0.82 & 0.82 & 0.82 & 0.82 & 0.42 & 0.80 & 0.55 & 0.92 & 0.74 & 0.82 \\
\midrule
\multirow{2}{*}{FR} 
& \makecell{Claude}      & \textbf{0.97} & \textbf{0.90} & \textbf{0.94} & \textbf{0.96} & \textbf{0.91} & \textbf{0.93} & \textbf{0.78} & \textbf{0.91} & \textbf{0.84} & \textbf{0.98} & \textbf{0.81} & \textbf{0.89} \\
& \makecell{Llama-based} & 0.93 & 0.74 & 0.83 & 0.92 & 0.73 & 0.81 & 0.77 & 0.74 & 0.76 & 0.95 & 0.65 & 0.77 \\
\midrule
\multirow{2}{*}{EL} 
& \makecell{Claude}      & 0.76 & \textbf{0.95} & 0.84 & 0.75 & \textbf{0.92} & 0.83 & 0.40 & \textbf{0.95} & 0.56 & 0.87 & \textbf{0.82} & \textbf{0.84} \\
& \makecell{Llama-based} & \textbf{0.83} & 0.90 & \textbf{0.86} & \textbf{0.82} & 0.86 & \textbf{0.84} & \textbf{0.45} & 0.90 & \textbf{0.60} & \textbf{0.91} & 0.71 & 0.80 \\
\bottomrule
\end{tabular}
}
\caption{Precision, recall, and F1-score across dataset variations, models, and languages. Bold values indicate the best score between the two models for each metric.}
\label{tab:precision}
\end{table*}

We evaluate the performance of our system with the 2 different LLM configurations across the three languages (English, French, and Greek) and the different variants of the annotated evaluation dataset.

\subsubsection{Overall Evaluation}
Table~\ref{tab:precision} reports precision, recall, and F1-score for each language, dataset variant, and LLM configuration. Overall, our results demonstrate the strong performance of the hybrid approach and reveal interesting patterns across models, languages, and dataset handling strategies.

\paragraph{Model Comparison} Across English and French, Claude consistently achieves higher precision, recall, and F1-score than the Llama-based models, highlighting its effectiveness on higher-resource languages. For Greek, the situation is more nuanced: the Llama-based KriKri model outperforms Claude in terms of precision, while Claude exhibits higher recall. Despite these differences, the overall F1-scores for the two models are comparable in Greek, differing by only 2--3\%, whereas in English and French the differences can reach up to 10\%.

\paragraph{Language Comparison} Focusing on overall performance (F1-score), the Llama-based models show comparable results across all three languages, indicating consistent behavior regardless of language. In contrast, Claude exhibits a notable drop in Greek, achieving F1-scores close to the much smaller Llama-based model. This difference is likely due to Claude’s reduced capabilities in Greek, which is a lower-resource language compared to English and French. The drop in Claude’s performance in Greek is mainly driven by lower precision. However, lower Precision in Greek is observed for both models compared to the other languages. This reduction in precision indicates that both models tend to over-flag texts as hateful in Greek, highlighting the difficulty of balancing sensitivity and specificity in a lower-resource language context.

\paragraph{Dataset Variant Comparison} Comparing the different strategies for handling borderline cases, we observe that the system performs better under the \emph{Strict} variant rather than \emph{Permissive}, indicating that it is sensitive in flagging texts as hate speech. However, this sensitivity does not lead to over-flagging, as performance on the \emph{Majority} variant, which is considered our primary gold standard, and the \emph{Safe} variant, which is more ``neutral'', is equally strong, if not slightly higher. This demonstrates that our system balances accurate detection with careful handling of uncertain cases.

\subsubsection{Pipeline Analysis}

\begin{table*}[ht]
\centering
\resizebox{0.9\textwidth}{!}{%
\begin{tabular}{ccc
>{\centering\arraybackslash}m{1.2cm}
>{\centering\arraybackslash}m{1.2cm}
>{\centering\arraybackslash}m{1.2cm}
>{\centering\arraybackslash}m{1.2cm}
>{\centering\arraybackslash}m{1.2cm}
>{\centering\arraybackslash}m{1.2cm}
>{\centering\arraybackslash}m{1.2cm}
>{\centering\arraybackslash}m{1.2cm}
>{\centering\arraybackslash}m{1.2cm}
>{\centering\arraybackslash}m{1.2cm}
>{\centering\arraybackslash}m{1.2cm}
>{\centering\arraybackslash}m{1.2cm}
}
\toprule
& & & \multicolumn{3}{c}{\makecell{Safe}} & \multicolumn{3}{c}{\makecell{Majority}} & \multicolumn{3}{c}{\makecell{Permissive}} & \multicolumn{3}{c}{\makecell{Strict}} \\
\cmidrule(lr){4-6} \cmidrule(lr){7-9} \cmidrule(lr){10-12} \cmidrule(lr){13-15}
Lang & Model & Type &
Precision & Recall & \mbox{F1-Score} & 
Precision & Recall & \mbox{F1-Score} & 
Precision & Recall & \mbox{F1-Score} & 
Precision & Recall & \mbox{F1-Score} \\
\midrule
\multirow{6}{*}{EN} 
% & \multirow{3}{*}{Claude}      & TF & 0.94 & 0.93 & 0.93 & 0.90 & 0.90 & 0.90 & 0.47 & 0.93 & 0.63 & 0.97 & 0.80 & 0.88 \\
& \multirow{3}{*}{Claude}      & TF & 0.99 & 0.83 & 0.90 & 0.93 & 0.76 & 0.84 & 0.52 & 0.83 & 0.64 & 0.99 & 0.67 & 0.80 \\
&                              & TB & 0.92 & 0.22 & 0.36 & 0.94 & 0.29 & 0.44 & 0.40 & 0.22 & 0.30 & 0.97 & 0.23 & 0.37 \\
&                              & F    & 0.97 & 0.90 & 0.93 & 0.92 & 0.89 & 0.90 & 0.47 & 0.90 & 0.62 & 0.98 & 0.78 & 0.87 \\
% \cmidrule[\lightrulewidth]{2-15}
\addlinespace[0.5em]
% & \multirow{3}{*}{Llama-based} & TF & 0.83 & 0.85 & 0.84 & 0.82 & 0.86 & 0.84 & 0.42 & 0.85 & 0.57 & 0.91 & 0.77 & 0.84 \\
& \multirow{3}{*}{Llama-based} & TF & 0.90 & 0.77 & 0.83 & 0.87 & 0.78 & 0.82 & 0.46 & 0.77 & 0.57 & 0.95 & 0.68 & 0.79 \\
&                              & TB & 0.80 & 0.30 & 0.44 & 0.81 & 0.36 & 0.50 & 0.36 & 0.30 & 0.33 & 0.91 & 0.32 & 0.48 \\
&                              & F  & 0.84 & 0.80 & 0.82 & 0.82 & 0.82 & 0.82 & 0.42 & 0.80 & 0.55 & 0.92 & 0.74 & 0.82 \\
\midrule
\multirow{6}{*}{FR} 
% & \multirow{3}{*}{Claude}      & TF & 0.97 & 0.91 & 0.94 & 0.96 & 0.91 & 0.93 & 0.77 & 0.91 & 0.83 & 0.98 & 0.83 & 0.90 \\
& \multirow{3}{*}{Claude}      & TF & 0.97 & 0.88 & 0.92 & 0.96 & 0.86 & 0.91 & 0.79 & 0.88 & 0.83 & 0.97 & 0.78 & 0.86 \\
&                              & TB & 1.00 & 0.29 & 0.45 & 1.00 & 0.29 & 0.45 & 0.82 & 0.29 & 0.43 & 1.00 & 0.26 & 0.41 \\
&                              & F  & 0.97 & 0.91 & 0.94 & 0.96 & 0.91 & 0.93 & 0.78 & 0.91 & 0.84 & 0.98 & 0.81 & 0.89 \\
% \cmidrule[\lightrulewidth]{2-15}
\addlinespace[0.5em]
% & \multirow{3}{*}{Llama-based} & TF & 0.90 & 0.69 & 0.78 & 0.88 & 0.69 & 0.78 & 0.71 & 0.69 & 0.70 & 0.92 & 0.64 & 0.75 \\
& \multirow{3}{*}{Llama-based} & TF & 0.94 & 0.55 & 0.70 & 0.93 & 0.53 & 0.67 & 0.79 & 0.55 & 0.65 & 0.95 & 0.48 & 0.63 \\
&                              & TB & 0.94 & 0.36 & 0.52 & 0.94 & 0.36 & 0.52 & 0.79 & 0.36 & 0.50 & 0.95 & 0.31 & 0.47 \\
&                              & F  & 0.93 & 0.74 & 0.83 & 0.92 & 0.73 & 0.81 & 0.77 & 0.74 & 0.76 & 0.95 & 0.65 & 0.77 \\
\midrule
\multirow{6}{*}{EL} 
% & \multirow{3}{*}{Claude}      & TF & 0.70 & 0.95 & 0.80 & 0.71 & 0.92 & 0.80 & 0.37 & 0.95 & 0.53 & 0.84 & 0.84 & 0.84 \\
& \multirow{3}{*}{Claude}      & TF & 0.78 & 0.92 & 0.85 & 0.76 & 0.87 & 0.81 & 0.42 & 0.92 & 0.58 & 0.88 & 0.75 & 0.81 \\
&                              & TB & 0.84 & 0.29 & 0.43 & 0.87 & 0.30 & 0.44 & 0.42 & 0.29 & 0.34 & 0.92 & 0.25 & 0.39 \\
&                              & F  & 0.76 & 0.95 & 0.84 & 0.75 & 0.92 & 0.83 & 0.40 & 0.95 & 0.56 & 0.87 & 0.82 & 0.84 \\
% \cmidrule[\lightrulewidth]{2-15}
\addlinespace[0.5em]
& \multirow{3}{*}{Llama-based} & TF & 0.88 & 0.80 & 0.84 & 0.85 & 0.75 & 0.80 & 0.50 & 0.80 & 0.62 & 0.93 & 0.59 & 0.72 \\
&                              & TB & 0.78 & 0.30 & 0.43 & 0.81 & 0.32 & 0.46 & 0.37 & 0.30 & 0.33 & 0.89 & 0.28 & 0.43 \\
&                              & F  & 0.83 & 0.90 & 0.86 & 0.82 & 0.86 & 0.84 & 0.45 & 0.90 & 0.60 & 0.91 & 0.71 & 0.80 \\
\bottomrule
\end{tabular}
}
\caption{Precision, recall, and F1-score of the Term Free (TF), Term Based (TB), and Fused (F) pipelines.}
\label{tab:pipelines-main}
\end{table*}

Table~\ref{tab:pipelines-main} presents the precision, recall, and F1-score of the two pipelines (term-based and term-free), along with the fused hybrid system, across model configurations, dataset variations and languages.
% Metrics for all dataset variants are provided in Appendix~\ref{app:pipeline-analysis}.

Across all languages and model configurations, the fused system consistently outperforms the individual pipelines in terms of overall performance (F1-score), showcasing that enhancing prompts with term descriptions from curated resources aids even large models, such as Claude Sonnet, despite their extensive implicit knowledge. The only exception is the English Llama-based model, where the fused system and the term-free pipeline obtain the same F1-score. This confirms the value of combining complementary signals from both pipelines.

As expected, the term-based pipeline shows high precision but substantially lower recall. This reflects its design: it captures only cases that involve inherently derogatory terms present in the curated vocabularies. Its contribution is therefore narrow but reliable. In contrast, the term-free pipeline covers a much broader space of hateful expressions and achieves higher recall.

We experimented with strengthening the term-free pipeline by explicitly prompting it to identify inherently derogatory terms in addition to broader hate speech. While this slightly improved its precision and recall, the term-free pipeline alone was still unable to reach the performance level of the fused system. This highlights the benefit of integrating the two complementary approaches rather than relying exclusively on a single LLM-based classifier.

Finally, it is important to note that the contribution of the term-based pipeline is not fully reflected in ``macroscopic'' evaluation metrics. Many commonly used slurs are already well covered by large language models, especially in high-resource languages, which limits the measurable impact of lexical resources when averaged across large datasets. However, the term-based pipeline plays a crucial role in capturing less frequent or newly emerging derogatory terms, which are underrepresented in existing datasets. It also strengthens the system’s robustness against evolving or intentionally obfuscated hateful expressions, an area where prompting-based approaches alone are more vulnerable. The hybrid design therefore, provides both better performance and better long-term adaptability.

\subsection{Quality of Explanations}

Beyond detection accuracy, evaluating the quality of the explanations produced by our system is a key aspect of our study. To this end, the 18 annotators who contributed to the evaluation dataset assessed the explanations generated by our system for the detections on the texts of the evaluation dataset. Each annotator reviewed explanations produced by both LLM configurations (Claude and the Llama-based models), presented in a mixed order.
The evaluators provided feedback along four dimensions: 
% --\emph{Explanation quality (1-5 scale)}: How well the explanation reflects why the text constitutes hate speech (relevance, completeness, correctness). \\
% --\emph{Explanation fluency (1-5 scale)}: How well-written the explanation is (grammar, syntax, readability). \\
% --\emph{Feedback on explanation}: ``Too vague''; ``Repetitive''; ``Incorrect Details''; ``Too Verbose''; ``Other''.
% --\emph{Additional comments} (optional free text). 
\begin{itemize}
    \item \emph{Explanation quality (5 point Likert scale)}: How well the explanation reflects why the text constitutes hate speech, considering relevance, completeness, and correctness. 
    \item \emph{Explanation fluency (5 point Likert scale)}: How well-written the explanation is, in terms of grammar, syntax, and readability.
    \item \emph{Feedback on explanation}: Categorical labels including ``Irrelevant Information'', ``Too vague'', ``Duplication of Information'', ``Incorrect Details'', ``Too Verbose'', and ``Other''.
    \item \emph{Additional comments}: Optional free-text field.
\end{itemize}

Table~\ref{tab:explanations} presents the average ratings and the most frequent issues (Issue 1 being the most frequent, and Issue 2 the 2nd most frequent) reported by annotators. Overall, explanations in English and French achieved high scores for both content and fluency (around 4 or higher), while Greek received lower scores. This reduction is likely due to a combination of factors: lower precision in Greek led to more false positives, which in turn caused the models to generate explanations that included irrelevant or incorrect information, lowering content quality. 

The two LLM configurations produced explanations of comparable quality. Claude generally scored slightly higher in both content and fluency, though differences were modest. Fluency ratings were relatively high across all languages, with a minor decrease in Greek, reflecting the models' reduced effectiveness in this lower-resource language, which aligns with our findings in the quantitative evaluation of Section~\ref{sec:accuracy} where we saw Claude's performance dropping significantly in Greek. Content scores showed larger variation, largely reflecting misclassifications: explanations generated to justify incorrect classifications naturally received lower content ratings, even if their fluency remained acceptable.

Analysis of the categorical feedback reveals that the most common issues across all languages were ``Irrelevant Info'' and ``Too Verbose''. These findings align with our observation that explanations for misclassified texts tended to include additional information to justify the model's decision, sometimes introducing irrelevant or excessive details. In some cases, this behavior reflects the model attempting to compensate for uncertainty, rather than outright hallucination. Free-text feedback also indicated that explanations occasionally exaggerated the level of toxicity in a text, which is consistent with our observation that the system is generally more eager to flag potential hate speech rather than under-flagging, while still maintaining good overall metrics.
For additional details regarding the feedback, including the distribution of reported issues across languages and models, we refer the reader to Appendix~\ref{app:exp-feedback}.

\begin{table}[t]
\centering
\resizebox{\columnwidth}{!}{
\begin{tabular}{cccccc}
\hline
\textbf{Lang} & \textbf{Model} & \makecell{Content} & \makecell{Fluency} & \makecell{Issue 1} & \makecell{Issue 2} \\
\hline
\multirow{2}{*}{EN} & Claude & 4.24 $\pm$ 1.27 & 4.66 $\pm$ 0.70 & \makecell{Irrelevant \\Info} & \makecell{Too Verbose } \\
& \makecell{Llama\\based} & 3.92 $\pm$ 1.43 & 4.59 $\pm$ 0.71 & \makecell{Irrelevant \\ Info} & \makecell{Other \\ } \\
\hline
\multirow{2}{*}{FR} & Claude & 4.45 $\pm$ 0.67 & 4.65 $\pm$ 0.61 & \makecell{Too Verbose} & \makecell{Irrelevant \\Info} \\
& \makecell{Llama\\based} & 4.40 $\pm$ 0.76 & 4.66 $\pm$ 0.59 & \makecell{Too Verbose} & \makecell{Irrelevant \\ Info} \\
\hline
\multirow{2}{*}{EL} & Claude & 3.35 $\pm$ 1.13 & 4.12 $\pm$ 0.60 & \makecell{Irrelevant \\ Info} & \makecell{Too Verbose} \\
& \makecell{Llama\\based} & 3.19 $\pm$ 1.14 & 4.04 $\pm$ 0.65 & \makecell{Irrelevant \\ Info} & \makecell{Too Verbose} \\
\hline
\end{tabular}
}
\caption{Explanations evaluation}
\label{tab:explanations}
\end{table}

\section{Conclusions}

% In this work, we present an approach that combines LLMs with curated in-domain knowledge and traditional NLP techniques for the detection of hate speech, along with the generation of high-quality contextual explanations for why certain expressions are considered contentious. The approach has been implemented for Greek, English, and French, and our evaluation demonstrates strong performance both in terms of detection accuracy and the quality of the generated explanations.
% The evaluation also offered valuable insights into the handling of borderline cases, revealing a tendency of the system to flag marginal expressions as positive. Furthermore, it highlighted the inherently subjective nature of hate speech assessment, with interpretations often influenced by factors such as ideological stances, demographics, and cultural biases~\cite{tonneau2025}.
% A key contribution of our work is also the development of three vocabularies of hate speech terms, enriched with information about their meaning and related identity-related categories. The vocabulary constitutes a valuable resource that can be exploited for the domain adaptation and improvement of similar initiatives.

In this work, we presented a hybrid system that combines LLMs with curated domain knowledge and curated vocabularies for detecting hate speech and generating contextual explanations across Greek, English, and French. Our evaluation demonstrates strong performance in both detection and explanation quality.
The evaluation also offered valuable insights into the handling of borderline cases, revealing a tendency of our system to flag marginal expressions as positive (hate speech), without, however, over-flagging, as our system achieved good metrics in the majority variant of our evaluation dataset, which is used as a gold standard. We also analyzed the contribution of the different pipelines, showing the benefit of our hybrid approach over LLM-only approaches, and the complementarity of the 2 pipelines. 
A key contribution of our work is the development of three vocabularies of inherently offensive terms, along with information about their meanings and related identity categories, a valuable resource for domain adaptation and similar initiatives. These vocabularies, along with the annotated evaluation dataset and the source code of the system are available at \url{https://github.com/ails-lab/detoex}.

We see two main directions for the further development of this work. The first is expanding coverage to additional languages by creating new vocabularies. The openly available vocabularies, construction process, and detection system also provide a foundation for the broader community to build upon, whether by developing systems in other languages, enriching further the existing ones, or finding new applications. The second direction is improving the system itself, where we intend to explore techniques such as Retrieval-Augmented Generation, self-evaluation, and consistency-based methods to enhance robustness and reliability.

% There are several directions for future work, both to improve the detection system’s performance and extend its applicability to additional languages. Building on the same methodology, vocabularies for more languages can be created, enabling the detection of hate speech in these languages with minimal adaptations to our system architecture and prompts. There is also significant room for exploring various fine-tuning and Retrieval-Augmented Generation approaches, as well as self-evaluation and consistency-based techniques that can enhance the system's robustness. %and faithfulness.
%A comparative evaluation with other existing hate speech detection systems that have similar objectives to ours is also left as future work. 
%Moreover, possible adaptations of the detection tool for new application areas, so that it can deal with other types of toxic language or biases, will also be considered.

%In the future, we aim to both improve the detection system’s performance and extend its applicability to additional languages. Building on the same methodology, we plan to create vocabularies for more languages, enabling the detection of hate speech in these languages with minimal adaptations to our system architecture and prompts. Additionally, we intend to explore recent advances in grounding methods, such as Retrieval-Augmented Generation (RAG), as well as self-evaluation and consistency-based techniques, to enhance the quality, faithfulness, and reliability of the generated explanations.

\section*{Limitations}

The following limitations of the current work should be acknowledged:
\begin{itemize}
    \item The preparation of the vocabularies of derogatory terms and corresponding categories and descriptions highly relied on an automatic process. In the case of Greek, all terms and associated information have been reviewed and improved, where necessary, by a human evaluator. For French and English, due to the high number of terms involved,  human review was limited to a 10\% sample of the vocabulary, with the results finding the automatically generated samples satisfactory in more than 90\% of the cases. Thus, we expect the vocabularies to be of high quality; however, it's still possible that they may include inaccurate information, and some descriptions may suffer from inferior quality. %This may in turn influence the performance of our system.
    \item The available ground truth data from previous initiatives does not fully align with the criteria following from the definition of hate speech we adopt. For Greek, the only dataset we were able to find includes annotations concerning toxic content, with only a subset of the positively labeled tweets actually being derogatory in relation to an identity characteristic. A similar remark holds for the French Hate Speech Superset, which also contains a significant number of aggressive language examples that do not constitute hate speech according to our definition. For English, we observe an opposite trend: a significant subset of the tweets drawn from the Hate Speech Superset is annotated as negative, even though they are clearly derogatory toward an identity characteristic according to our definition. 
    % For instance, references to ``kung flu'' are often annotated as negative in the Hate Speech Superset, whereas they would be considered positive cases (i.e., hate speech) under our definition. 
    These discrepancies may stem from annotation errors or other types of inconsistencies, such as those described in ~\cite{vidgen2020}, or may reflect differences in how hate speech is interpreted. For these reasons, we were unable to use the labels of the datasets as ground truth for our evaluation, and we had to rely solely on human input to accurately calculate the performance of our system. However, the selection of annotators who have experience with the topic of hate speech gives us confidence in the quality of the annotations, giving as an high-quality evaluation dataset. 
    \item Conducting a fair and reliable comparative evaluation with other hate speech detection systems is particularly challenging. As discussed above, issues related to ground truth data limit the availability of suitable benchmark datasets, especially for Greek and French. Moreover, existing systems vary widely in scope. As identified in previous work~\cite{Davidson2017}, many systems do not distinguish between generally offensive language (e.g., personal insults, individual bullying), which our system treats as negative examples, and hate speech, which targets groups defined by identity characteristics. At the same time, several systems~\cite{vidgen2020,chiril2020} focus only on a subset of identity characteristics (e.g., gender). Owing to these limitations, we rely on human validation for the evaluation presented in this work. 
    % The design of a robust comparative methodology—one that accounts for both appropriate baselines and the specific features of different detection systems—requires dedicated effort and is left for future work.
\end{itemize}

\section*{Ethical Considerations}
Before conducting the human evaluation, all participants were thoroughly briefed on the task through an introductory workshop organized by the authors. During this session, we explained the evaluation procedure step-by-step, demonstrated example cases, and addressed questions to ensure that all evaluators clearly understood the objectives and evaluation criteria. The session was recorded and made available to participants for future reference, and a dedicated communication channel remained open throughout the evaluation period for any additional clarification or support.

Given the potentially disturbing nature of the content, participants were warned about the presence of offensive or harmful language before beginning the task. The study was designed and supervised by experts in abusive language and hate speech research, who have been actively involved in education and initiatives related to harmful language. These experts also facilitated the workshop and were available for ongoing consultation.

Participants were informed that their participation was voluntary and that they could withdraw from the study at any time without consequence. While the experts involved in evaluating the vocabularies, and designing and overseeing the human evaluation were properly compensated, the evaluators themselves participated on a voluntary basis, due to limited funding availability.

%DETOEX made use of established pretrained LLMs -Claude by Anthropic and LLM versions based on Llama 3.1-8B, which are open source. Both models align with the European AI Act and are compliant with privacy regulations, while adopting mechanisms that filter out and correct known sources of misinformation and strategies to avoid biases in trained data. All data used to ground and guide the LLM were collected in accordance with fairness and data minimisation principles, guaranteeing that all data processed throughout DETOEX is relevant and limited to the purposes of the research project. Regarding the human evaluation process, participation was voluntary. Concerning the protection of personal data, we ensured that appropriate GDPR-compliant procedures were in place. Among others, the following were observed: informed consent; measures ensuring confidentiality, concerning the collection, storage and management of data; and the right to withdraw.

 \section*{Acknowledgments}
The work presented in this paper has been co-funded by the European Commission under the project "DEtection of TOxic and hateful speech with EXplanations -DETOEX" (UTTER - Unified Transcription and Translation for Extended Reality Agreement No. 101070631- HE; Grant Agreement No. 10039436 - UKRI; FSTP).
% The work presented herein has been funded by the European Commission as part of a Financial Support for Third Parties with the UTTER (Unified Transcription and Translation for Extended Reality) project.
We would also like to express our sincere gratitude to all the evaluators who participated in the DETOEX evaluation process. Their valuable contributions were instrumental to this work and greatly appreciated.
% lorem ipsum

\bibliography{custom}

\appendix
% \onecolumn

\section{Vocabulary Creation}
\label{app:voc_example}

We curated three vocabularies (one per the considered languages) that contain derogatory and offensive terms, which have at least one meaning that can be considered hate speech, i.e. terms that are per se derogatory and offensive towards a group and their members, based on their characteristics or beliefs. In this respect, the vocabularies were not meant to include common slurs, which may be directed against certain groups considered by our categorization in certain contexts, but are not by themselves derogatory or offensive towards such a group (e.g., ``asshole'').

The vocabularies contain the following pieces of information:

\begin{itemize}
    \item Term: the term that can be used as hate speech
    \item Description: Free-text description of all the meanings of the term (including both offensive and non-offensive ones). The description should mention why and under which circumstances the term is used in an offensive way.
    \item Category: The group or groups towards which the term is by itself offensive or derogatory, based on the groups defined in Section \ref{sec:aspects}.
    \item Source: A link to the repository from which the term was sourced.
\end{itemize}

See Table \ref{tab:voc_example} for an example.

\begin{table*}[ht]
    
    \centering
    \begin{tabular}{>{\RaggedRight\arraybackslash}p{0.08\textwidth}
>{\RaggedRight\arraybackslash}p{0.62\textwidth}>{\RaggedRight\arraybackslash}p{0.15\textwidth}>{\RaggedRight\arraybackslash}p{0.15\textwidth}}
        \toprule
        Term & Description & Categories & Source\\
        \midrule
        bitch & The term 'bitch' is primarily offensive when used to refer to women in a derogatory manner, implying they are aggressive, unpleasant, or overly assertive—traits that would often be viewed positively in men. It's also problematic when applied to men to suggest weakness or effeminacy, as this usage reinforces harmful gender stereotypes by equating femininity with inferiority. While the word has a neutral meaning when referring to female dogs, its use as a slur has overshadowed this definition in most contexts. In some LGBTQ+ communities and among close friends, the term has been reclaimed and may be used affectionately, but this usage is context-dependent and generally inappropriate for those outside these communities. The term's evolution from canine terminology to gendered insult reflects long-standing societal attitudes that devalue women and feminine characteristics. & Gender; Sexual Orientation & \url{https://en.wiktionary.org/wiki/bitch}\\
        \bottomrule
    \end{tabular}
    \caption{A sample entry from the vocabulary.}
    \label{tab:voc_example}
\end{table*}

We used the English, French and Greek Wiktionaries to create the respective vocabularies. This decision was primarily motivated by a lack of openly-available and extensive enough alternatives, especially in Greek.

In order to fetch an initial set of term candidates we made use of Wiktionary categories and tags that denoted a derogatory or insulting usage of a term. These categories and tags were:

\begin{itemize}
    \item English
    \begin{enumerate}
        \item Category:English derogatory terms
        \item  Category:English vulgarities
        \item Category:English offensive terms
    \end{enumerate}
    \item French
    \begin{enumerate}
        \item Catégorie:Termes péjoratifs en français
        \item Catégorie:Insultes en français
    \end{enumerate}
    \item Greek
    \begin{enumerate}
        \item \textgreek{Μειωτικοί όροι (νέα ελληνικά) | μειωτικός | μειωτική | μειωτικό | μειωτικά}
        \item \textgreek{Κατηγορία:Υβριστικοί όροι (νέα ελληνικά) | υβριστικός | υβριστική | υβριστικό | υβρισιτκά}
        \item \textgreek{Κατηγορία: Χυδαιολογίες (νέα ελληνικά) | χυδαίος | χυδαία | χυδαίο}
        \item \textgreek{βρισιά | βρισιές}
    \end{enumerate}
\end{itemize}

The ``query'' action of the Wiktionary API was used to fetch pages. This is an example of an API call that fetches all terms under the category  ``\textgreek{Μειωτικοί όροι (νέα ελληνικά)}'' (derogatory terms in Greek):

\begin{tcolorbox}
\ttfamily\small
\textbf{GET} \texttt{https://el.wiktionary.org/w/api.php}

\textbf{Request Body:}

\begin{lstlisting}[language=json]
{
    "action": "query",
    "list": "categorymembers",
    "cmtitle": "(*\textgreek{Κατηγορία: Μειωτικοί\_όροι\_(νέα\_ελληνικά)}*)",
    "cmprop": "title|ids",
    "format": "json",
    "cmlimit": "500"
}
\end{lstlisting}
\end{tcolorbox}

This is an example of an API call that fetches all terms tagged as ``\textgreek{υβριστικός}'' (insulting):

\begin{tcolorbox}
\ttfamily\small
\textbf{GET} \texttt{https://el.wiktionary.org/w/api.php}

\textbf{Request Body:}

\begin{lstlisting}[language=json]
{
    "action": "query",
    "prop": "linkhere",
    "cmtitle": "(*\textgreek{υβριστικός}*)",
    "cmprop": "title|ids",
    "format": "json",
    "lhlimit": "500"
}
\end{lstlisting}
\end{tcolorbox}

This resulted in 11,310 English, 3,749 French and 965 Greek term candidates.
Definitions were extracted from the term pages by parsing the HTML and detecting the sections containing definitions. Since the Wiktionary pages are only designed to be human-readable there is no consistent way to isolate the sections containing definitions, so some assumptions and work-arounds had to be made. Sections containing definitions almost universally start with a heading describing the part-of-speech of the word. A page may contain many such sections containing definitions (e.g. the term ``\textgreek{αδερφή}'' contains definitions both under section ``\textgreek{Ουσιαστικό}'' and  ``\textgreek{Κλιτικός τύπος επιθέτου}''). All heading tags for all pages were collected and the ones referring to parts-of-speech were isolated. Other sections containing irrelevant information, such as the pronunciation of a term, were discarded.  Within the sections kept, the term definitions are almost exclusively structured as a list (even if the section contains a single definition), so the HTML tags <ul> and <ol> were used to detect the lists of term definitions. For each definition, only plain text was kept, discarding any links or tags contained in the HTML. For each term, all definitions collected were unified in a single numbered list and stored in a CSV file.

The next step was only keeping the terms that did actually have some derogatory usage and could be used as hate speech. Ideally, we would like to manually review all fetched terms. Although this proved feasible in Greek, with the help of some native speakers, it was not possible for English or French due to the sheer amound of the terms fetched. Instead, we opted for using Claude Sonnet 3.7 to filter the terms. Claude Sonnet 3.7 was also used to create the vocabulary description of each term by fusing the different Wiktionary definitions of each term into a single, coherent piece of text that describes all usages of the term, noting which ones are derogatory. See Appendix \ref{app:prompts} for the prompt used for filtering and description creation. Claude's performance on filtering terms for English and French and constructing descriptions in all three languages was evaluated by humans. Evaluators agreed with over 90\% of Claude's filtering and found the descriptions of sufficient quality.

A limitation that we inspected in some descriptions is that Wiktionary does not always sufficiently reflect probable recent shifts in the usage of terms in certain social environments, especially regarding the extent to which certain terms are used in a reclaimed manner (e.g. the reclaiming of ``tranny'' by trans people). As a result, the descriptions generated by Claude based on the Wiktionary definitions often treat such reclaimed usages as being derogatory. Where possible, this limitation has been addressed by adapting the automatic descriptions accordingly. Furthermore, the LLMs used for the tool's pipeline were also prompted to be cautious of such usages by the affected community itself, so as to avoid over-flagging (see Appendix \ref{app:prompts} for the relevant prompts).

\section{Vocabulary Categories}
\label{app:vocab-stat}

% \begin{table}[ht]
% \centering
% \begin{tabular}{lrrr}
% \toprule
% Category & English & French & Greek \\
% \midrule
% Addiction & 38 & 20 & 8 \\
% Age & 85 & 57 & 28 \\
% Disability & 266 & 63 & 17 \\
% Ethnicity & 1194 & 449 & 45 \\
% Gender & 1109 & 410 & 61 \\
% Physical Appearance & 201 & 80 & 26 \\
% Political Affiliation & 774 & 382 & 28 \\
% Public Institutions & 261 & 148 & 13 \\
% Race & 842 & 189 & 12 \\
% Religion & 400 & 177 & 10 \\
% Sexual Orientation & 455 & 102 & 47 \\
% Socioeconomic & 335 & 279 & 31 \\
% Other & 11 & 7 & 23 \\
% \bottomrule
% \end{tabular}
% \caption{Counts of category tags for vocabulary terms by language}
% \label{tab:category_counts}
% \end{table}

\begin{table*}[ht]
\centering
\begin{tabular}{lrrr|r}
\toprule
Category & English & French & Greek & Total \\
\midrule
Addiction & 38 & 20 & 8 & 66 \\
Age & 85 & 57 & 28 & 170 \\
Disability & 266 & 63 & 17 & 346 \\
Ethnicity & 1194 & 449 & 45 & 1688 \\
Gender & 1109 & 410 & 61 & 1580 \\
Physical Appearance & 201 & 80 & 26 & 307 \\
Political Affiliation & 774 & 382 & 28 & 1184 \\
Public Institutions & 261 & 148 & 13 & 422 \\
Race & 842 & 189 & 12 & 1043 \\
Religion & 400 & 177 & 10 & 587 \\
Sexual Orientation & 455 & 102 & 47 & 604 \\
Socioeconomic & 335 & 279 & 31 & 645 \\
Other & 11 & 7 & 23 & 41 \\
% \midrule
% \textbf{Total} & \textbf{6971} & \textbf{2363} & \textbf{350} & \textbf{9684} \\
\bottomrule
\end{tabular}
\caption{Counts of category tags for vocabulary terms by language}
\label{tab:category_counts}
\end{table*}

In table~\ref{tab:category_counts} we present the number of terms with each category label in our vocabularies. 
Note that each term can have multiple category tags, hence the sum of each column is greater than the number of terms in the respective language. 

\section{Prompts}
\label{app:prompts}

% In Table \ref{tab:prompts} 
Below we present the English prompts for all parts of the pipeline (term-based detection, term-free detection, and fusion of explanations), as well as vocabulary creation. The prompts for other languages can be found in the supplementary material and follow the exact same format, with the text having been translated by proficient speakers of the respective language.

Aspects of the prompts were adapted by manual evaluation of small samples, but without thorough benchmarking of prompting strategies and variations due to budgetary constraints and the different focus of this work. These are some examples of interest: The initial Term-Based prompt tended to flag all occurrences of a term as harmful, even when it contextually carried a different meaning. Chain of thought reasoning was added to ensure that the term has been properly disambiguated before flagging the text, which seemed to improve results. Both the Term-Based and Term-Free pipelines were seen to flag quotations in several cases, one in particular being news headlines, so the explicit instruction to not flag quotations or indirect speech was added and it seemed to eliminate this error.

%------------------------
\begin{tcolorbox}[colback=gray!5!white, colframe=black!75!black, title=Term-based detection - System Prompt, fonttitle=\bfseries, sharp corners=south, breakable]
You are an expert content moderator specializing in detecting hate speech in text. Your task is to analyze text and distinguish hateful from neutral uses of a specific term based on the following definition:

\textbf{Hate speech} refers to spoken or written communication that attacks or uses pejorative or discriminatory language with reference to a person or a group based on identity-related characteristics, including: Gender, Sexual orientation, Race, Ethnicity, Religion, Political affiliation, Socioeconomic status, Occupation, Age, Disability, Addiction, Physical appearance.

\textbf{You will be given:}
\begin{itemize}
    \item A term
    \item A description of how that term can be used in hateful and non-hateful/neutral ways
    \item A piece of text containing this term
    \item One or more target characteristics the term may be offensive toward (e.g., "Sexual Orientation", "Ethnicity")
\end{itemize}

Your goal is to analyze the text and then decide if the term is used as hate speech. Use the following reasoning steps:
\begin{enumerate}
    \item \textbf{Step 1}: If the description includes multiple possible meanings of the term, identify which meaning is used in the text. If disambiguation is particularly difficult, rely on non-hateful uses of the term. If it has only one clear meaning, write "Non ambiguous term". Do not evaluate the presence of hate speech yet.
    \item \textbf{Step 2}: Based on the meaning you identified, consider whether the term corresponds to the hateful usage described earlier. Consider both the possibility of it being used in a hateful way and the possibility of it being used in a neutral/non hateful way.
    \item \textbf{Step 3}: Decide whether the use of the term in the text is hateful or not and simply write "Hateful" or "Non hateful".
    \item \textbf{Step 4}: Provide a clear, concise explanation (under 100 words) of your judgment. In your explanation use the phrasing provided in the term description you will be given. Do not include, or refer to any previous Step.
\end{enumerate}

\textbf{Important considerations for analysis:}
\begin{itemize}
    \item \textbf{Indirect speech}: Any hate speech contained in the text as part of a quote or paraphrased from a different source should influence your decision significantly less or not at all.
    \item \textbf{Reclaimed language}: Some terms which are usually derogatory can be used in a reclaimed, empowering way by members of the same community they target. In these cases, the level of hate speech should be significantly lower or non-existent.
    \item \textbf{Self-identity versus targeting others}: Distinguish between someone who describes themself or their own community versus targeting others with the same language.
\end{itemize}

Format your output using XML tags as follows:

\texttt{<STEP\_1>} [Step 1] \texttt{</STEP\_1>} \\
\texttt{<STEP\_2>} [Step 2] \texttt{</STEP\_2>} \\
\texttt{<STEP\_3>} [Step 3] \texttt{</STEP\_3>} \\
\texttt{<STEP\_4>} [Step 4] \texttt{</STEP\_4>}
\end{tcolorbox}

%------------------------
\begin{tcolorbox}[colback=gray!5!white, colframe=black!75!black, title=Term-based detection - User Prompt, fonttitle=\bfseries, sharp corners=south]
\textbf{Term:} \{\} \\
\textbf{Description:} \{\} \\
\textbf{Text:} \{\} \\
\textbf{Characteristics:} \{\}
\end{tcolorbox}

%------------------------
\begin{tcolorbox}[colback=gray!5!white, colframe=black!75!black, title=Non-term-based detection - System Prompt, fonttitle=\bfseries, sharp corners=south, breakable]
You are an expert content moderator specializing in detecting hate speech in text. Your task is to analyze text and determine if it contains hate speech based on the following definition:

\textbf{Hate speech} refers to spoken or written communication that attacks or uses pejorative or discriminatory language with reference to a person or a group based on identity-related characteristics, including: Gender, Sexual orientation, Race, Ethnicity, Religion, Political affiliation, Socioeconomic status, Occupation, Age, Disability, Addiction, Physical appearance.

\textbf{Important considerations for analysis:}
\begin{itemize}
    \item \textbf{Non-targeted speech}: Do not consider slurs as hate speech, unless they are directed towards an individual or group defined by their identity characteristics.
    \item \textbf{Indirect speech}: Any hate speech that the text contains as part of a quote or paraphrased from a different source should affect your decision significantly less or not at all.
    \item \textbf{Self-identity versus targeting others}: Distinguish between someone who describes themself or their own community versus targeting others with the same language.
\end{itemize}

\textbf{Output Requirements:} Provide a decision and concise explanation (under 100 words) covering:
\begin{itemize}
    \item Which elements influenced your decision
    \item If hate speech was found, which identity characteristics are targeted
    \item Any ambiguities or nuances you considered
    \item Specific quotes from the text when necessary for your argument
\end{itemize}

Always use the exact XML format specified in the user prompt.
\end{tcolorbox}

%------------------------
\begin{tcolorbox}[colback=gray!5!white, colframe=black!75!black, title=Non-term-based detection - User Prompt, fonttitle=\bfseries, sharp corners=south, breakable]
Analyze the following text and evaluate whether it contains hate speech:

\textbf{Text for analysis:} \{\}

\textbf{Please give your response exactly as specified in the following format:}

\texttt{<DECISION>} ["Hate speech" or "Not hate speech"] \texttt{</DECISION>}

\texttt{<EXPLANATION>} [Explanation of your evaluation] \texttt{</EXPLANATION>}
\end{tcolorbox}

%------------------------
\begin{tcolorbox}[colback=gray!5!white, colframe=black!75!black, title=Fusion - System Prompt, fonttitle=\bfseries, sharp corners=south, breakable]
\textbf{Task: Merge Hate Speech Analysis Texts}

You will be given two or more separate texts describing hate speech content. Merge these analyses into a single, coherent description without redundancy.

\textbf{Instructions:}
\begin{enumerate}
    \item Combine the information from all texts into a unified analysis
    \item Reuse the existing text
    \item Remove duplicate information
    \item Reorganize for better flow
    \item Maintain accuracy
    \item Keep it focused
    \item Keep it brief
\end{enumerate}

\textbf{Input Format:} Text 1, Text 2, etc. \\
\textbf{Output Format:} Provide a single well-structured paragraph without opening/closing remarks.

\textbf{Example:} 
\begin{itemize}
    \item \textbf{Text 1:} The term "bitch" in this tweet is used as hate speech as it is part of a gender-based slur. The phrase aims to diminish and demean a woman through sexist language, linking her to derogatory references to sexual behavior and gendered stereotypes. The use of the term in this context violates basic principles of respect and gender equality.
    \item \textbf{Text 2:} The text contains hate speech that targets individuals based on their religion. Specifically, the term "diaperhead" is a derogatory and dehumanizing slur for Muslims, mocking traditional headwear like turbans or keffiyehs by comparing them to diapers. In addition, the text contains sexist language ("you stupid bitch") that targets the gender of the recipient and includes sexual insinuations of violence ("waiting to be fucked"). The language is directly offensive and targeted, without being a quotation or indirect speech.
    \item \textbf{Merged Output:} This particular text presents multiple levels of hate speech. The term "bitch" in this tweet is part of a gender-based slur. The phrase aims to diminish and humiliate a woman through sexist language and includes sexual innuendos of violence ("waiting to be fucked"). It is also a derogatory and dehumanizing characterization of Muslims, using the word "diaperhead", mocking traditional headwear like turbans or keffiyehs by comparing them to diapers.
\end{itemize}
\end{tcolorbox}

%------------------------
\begin{tcolorbox}[colback=gray!5!white, colframe=black!75!black, title=Fusion - User Prompt, fonttitle=\bfseries, sharp corners=south]
Please merge the following hate speech analyses: \{\}
\end{tcolorbox}

%------------------------
\begin{tcolorbox}[colback=gray!5!white, colframe=black!75!black, title=Vocabulary Creation, fonttitle=\bfseries, sharp corners=south, breakable]
You have an expert understanding of the English language and slang, and how it can be used in a derogatory manner to target individuals or groups through stereotypes, negative generalizations, or the use of identity-related markers (e.g., ethnicity, origin, profession) as a negative trait. This derogatory nature may be evident in the etymology or structure of the word (e.g., compound words using a component metaphorically to evoke a stereotype).

Your task is to help the user create a vocabulary of \textbf{English terms that can constitute hate speech}. You will be given a term and one or more short descriptions (e.g., definitions or usage contexts) extracted from the English Wiktionary. Not all terms you will be given constitute hate speech. Your task is to determine:

1. Whether the term can constitute hate speech, based on the following definition: "Hate speech refers to spoken or written communication that attacks or uses pejorative or discriminatory language with reference to a person or a group based on identity-related characteristics. These characteristics include: gender, sexual orientation, race, ethnicity, religion, political affiliation, socioeconomic status, occupation, age, disability, addiction, and physical appearance." Words must be offensive due to their \textbf{meaning} and \textbf{harmful to a group} with shared identity characteristics. Inherently degrading expressions qualify. Words that are merely offensive on an individual level, like common slurs, but not harmful to a social group, do \textbf{not} qualify. Some terms can have \textbf{neutral or non-offensive meanings} in other contexts. You should base your judgment on the provided description(s). If the description suggests the word is used in a derogatory sense targeting a group identity, it \textbf{can} constitute hate speech.

2. If the term \textbf{can} constitute hate speech, indicate which \textbf{category or categories} it targets, choosing from: Gender, Sexual orientation, Race, Ethnicity, Religion, Political affiliation, Socioeconomic status, Occupation, Age, Disability, Addiction, Physical appearance.

3. If the term \textbf{can} constitute hate speech, provide a vocabulary entry with clear, concise description for the term that explains:
\begin{itemize}
    \item In which context(s) the term is considered offensive or inappropriate
    \item If and when the term can be used in a neutral or acceptable way
    \item Why or how the term came to acquire its derogatory meaning, if such information is available
    \item Any relevant linguistic, cultural, or historical background
\end{itemize}

Output your answer as JSON using the following fields:
\begin{itemize}
    \item "reasoning": String with a short explanation of your decision in freeform text
    \item "hate\_speech": Boolean (true or false)
    \item "categories": Array with one or more categories, or empty if false
    \item "vocabulary\_entry": String, empty if false
\end{itemize}

\textbf{Examples:}

\textbf{EXAMPLE 1} \\
\textbf{INPUT:} TERM: welfare queen \\
DESCRIPTION(S): A woman collecting welfare, seen as doing so out of laziness, rather than genuine need. \\
\textbf{OUTPUT:} 
\{
"reasoning": "The term \textbackslash"welfare queen\textbackslash" is toxic because it's rooted in racist and sexist stereotypes that falsely portray people receiving government assistance as fraudulent and lazy, particularly targeting Black women.",
"hate\_speech": true,
"categories": ["Race", "Socioeconomic Status"],
"vocabulary\_entry": "The term \textbackslash"welfare queen\textbackslash" refers to someone who allegedly exploits government welfare systems, but it's primarily considered derogatory and offensive. Popularized during Ronald Reagan's 1976 presidential campaign, the phrase has been widely criticized for perpetuating harmful stereotypes about welfare recipients, particularly women of color, and for reinforcing classist and racist assumptions about poverty. The term is inappropriate in respectful discourse about social policy, though it may appear in academic or analytical contexts when discussing political rhetoric or media representation, provided its problematic history is acknowledged."
\}

\textbf{EXAMPLE 2} \\
\textbf{INPUT:} TERM: nerd \\
DESCRIPTION(S):
\begin{itemize}
    \item A person who is intellectual but generally introverted.
    \item One who has an intense, obsessive interest in something.
    \item A member of a subculture revolving around intellectualism, technology, video games, fantasy and science fiction, comic books and assorted media.
    \item One who is socially inept or unattractive, regardless of their intelligence; a social outcast.
\end{itemize}
\textbf{OUTPUT:}
\{
"reasoning": "While potentially teasing, 'nerd' does not target an identity group based on protected characteristics.",
"hate\_speech": false,
"categories": [],
"vocabulary\_entry": ""
\}

\textbf{EXAMPLE 3} \\
\textbf{INPUT:} TERM: slut \\
DESCRIPTION(S):
\begin{itemize}
    \item A sexually promiscuous woman.
    \item A prostitute.
    \item Any sexually promiscuous person.
    \item Someone who seeks attention through inappropriate means or to an excessive degree.
    \item A disloyal individual; someone who does not commit to a particular thing.
    \item A slovenly, untidy person, usually a woman.
    \item A bold, outspoken woman.
    \item A female dog.
    \item A maidservant.
    \item A rag soaked in a flammable substance and lit for illumination.
    \item To wear slutty clothing or makeup, or otherwise behave in a slutty manner.
    \item To visit places frequented by men, with the intention of engaging in sexual intercourse by means of flirting.
\end{itemize}
\textbf{OUTPUT:}
\{
"Reasoning": "The term slut has been used to stigmatized people for their sexual practices, especially women.",
"hate\_speech": true,
"categories": ["Gender", "Sexual Orientation"],
"vocabulary\_entry": "The term \textbackslash"slut\textbackslash" is a derogatory word traditionally used to shame women for perceived sexual promiscuity or non-conformity to conventional sexual norms. Historically rooted in misogynistic attitudes that police women's sexuality while applying different standards to men, the term is considered offensive and inappropriate in most contexts because it perpetuates harmful double standards and slut-shaming. However, the word has undergone some reclamation efforts, particularly in feminist and LGBTQ+ communities, where individuals may use it self-referentially or positively to challenge sexual stigma—though this reclaimed usage should only be employed by those within these communities and with clear understanding of the context."
\}
\end{tcolorbox}

\section{Inter-rater Agreement}
\label{app:inter-rater}
Inter-rater agreement was measured using Krippendorff’s alpha \cite{krippendorff2018content}, interpreting the three labels as interval values (0.0 for No, 0.5 for Unsure, 1.0 for Yes). Table~\ref{tab:inter_rater_agreement} provides an overview of the agreement scores, along with the percentage of texts with strict disagreement (at least one `\textit{Yes}' and one `\textit{No}'). 

\begin{table}[h]
\centering
\resizebox{\columnwidth}{!}{%
\begin{tabular}{lcc}
\toprule
& \makecell{Krippendorff's\\alpha} & 
\makecell{Percentage of tweets with\\disagreement between the raters} \\
\midrule
EN & 0.54 & 0.29 \\
FR & 0.74 & 0.10 \\
EL & 0.65 & 0.22 \\
\bottomrule
\end{tabular}
}
\caption{Inter-rater agreement}
\label{tab:inter_rater_agreement}
\end{table}

\section{Dataset Statistics}
\label{app:dataset-statistics}
In Table~\ref{tab:dataset-class-balance} we present the percentage of positive (containing hate speech) samples in each variant of the evaluation dataset in each language. We see that the classes are relatively balanced, especially for the majority label.

\begin{table}[h]
\centering
\begin{tabular}{lrrrr}
\toprule
& Safe & Majority & Permissive & Strict \\
\midrule
EN & 43\% & 50\% & 27\% & 64\% \\
FR  & 52\% & 54\% & 43\% & 60\% \\
EL   & 32\% & 38\% & 21\% & 55\% \\
\bottomrule
\end{tabular}
\caption{Percentage of texts labeled as positive across dataset variations and languages}
\label{tab:dataset-class-balance}
\end{table}

Table \ref{tab:ground-truth-agreement} shows the agreement between the labels that resulted from our annotation campaign and the labels of the original datasets. Agreement is generally low, especially for French. The English and French datasets seem to employ a more narrow definition of what constitutes hate speech, presenting higher agreement with the permissive labels. On the other hand, the Greek dataset flags more tweets as positive, which was expected as it is an offensive speech dataset rather than a hate speech one. This validates our choice of conducting annotation campaigns to acquire labels closer to our definition of hate speech, rather than rely on the original labels of the datasets.

\begin{table}[h]
\centering
\begin{tabular}{lrrrr}
\toprule
& Safe & Majority & Permissive & Strict \\
\midrule
EN & 71\% & 66\% & 74\% & 53\% \\
FR  & 58\% & 57\% & 58\% & 54\% \\
EL   & 76\% & 72\% & 62\% & 72\% \\
\bottomrule
\end{tabular}
\caption{Agreement rate between annotator labels and the original dataset labels}
\label{tab:ground-truth-agreement}
\end{table}

\section{Explanation Feedback}
\label{app:exp-feedback}

In Table~\ref{tab:error_categories}, we report the percentages of issues noted by evaluators for explanations with lower content or fluency scores. Importantly, the feedback field was filled in only for these lower-scoring explanations, which occurred infrequently. The percentages in the table are calculated only over the explanations for which evaluators provided feedback, so each column sums to 100\%, and they should not be interpreted as the proportion of all explanations in the dataset.

% \begin{table*}[ht]
% \centering
% \begin{tabular}{lrrrrrr}
% \toprule
% & \multicolumn{2}{c}{English} & \multicolumn{2}{c}{French} & \multicolumn{2}{c}{Greek} \\
% \cmidrule(lr){2-3} \cmidrule(lr){4-5} \cmidrule(lr){6-7}
% Category & Claude & Llama & Claude & Llama & Claude & Llama \\
% \midrule
% Duplication of Information & 3 & 2 & 1 & 1 & 8 & 4 \\
% Incorrect Details & -- & -- & -- & -- & 1 & 1 \\
% Irrelevant Information & 31 & 50 & 3 & 2 & 33 & 35 \\
% Too Vague & -- & -- & -- & -- & 1 & 2 \\
% Too Verbose & 14 & 9 & 13 & 4 & 12 & 15 \\
% Other & 10 & 11 & -- & 1 & 5 & 4 \\
% \bottomrule
% \end{tabular}
% \caption{Issues reported for the explanations accompanying hate speech detections per language and LLM}
% \label{tab:error_categories}
% \end{table*}

\begin{table*}[ht]
\centering
\begin{tabular}{lrrrrrr}
\toprule
& \multicolumn{2}{c}{English} & \multicolumn{2}{c}{French} & \multicolumn{2}{c}{Greek} \\
\cmidrule(lr){2-3} \cmidrule(lr){4-5} \cmidrule(lr){6-7}
Category & Claude & Llama & Claude & Llama & Claude & Llama \\
\midrule
Duplication of Information & 5.17\% & 2.78\% & 5.88\% & 12.50\% & 13.33\% & 6.56\% \\
Incorrect Details & -- & -- & -- & -- & 1.67\% & 1.64\% \\
Irrelevant Information & \textbf{53.45}\% & \textbf{69.44}\% & 17.65\% & 25.00\% & \textbf{55.00}\% & \textbf{57.38}\% \\
Too Vague & -- & -- & -- & -- & 1.67\% & 3.28\% \\
Too Verbose & 24.14\% & 12.50\% & \textbf{76.47}\% & \textbf{50.00}\% & 20.00\% & 24.59\% \\
Other & 17.24\% & 15.28\% & -- & 12.50\% & 8.33\% & 6.56\% \\
\bottomrule
\end{tabular}
\caption{Issues reported for the explanations accompanying hate speech detections per language and LLM}
\label{tab:error_categories}
\end{table*}

\section{Hardware}
\label{app:hardware}
All inference using Claude was performed via Amazon Web Services. Inference using the Llama based models was done on a server with a NVIDIA GeForce RTX 4090 GPU. All other code was executed locally in a single-threaded setup on a laptop with an AMD Ryzen 7 5800H CPU and 16 GB of RAM.

\end{document}